\documentclass{article}

    \usepackage[final]{neurips_2022}

\usepackage[utf8]{inputenc} 
\usepackage[T1]{fontenc}    
\usepackage{hyperref}       
\usepackage{url}            
\usepackage{booktabs}       
\usepackage{amsfonts}       
\usepackage{nicefrac}       
\usepackage{microtype}      
\usepackage{xcolor}         
\usepackage{amsmath,amssymb,amsfonts}
\usepackage{algorithmic}
\usepackage{graphicx}
\usepackage{textcomp}
\usepackage{xcolor}
\usepackage{array}
\usepackage{caption}
\usepackage{subcaption}

\title{Improving Assistive Robotics with Deep Reinforcement Learning}

\author{%
  Yash Jakhotiya\thanks{Corresponding author}\\
  Georgia Institute of Technology\\
  \texttt{yashjakhotiya@gatech.edu} \\
   \And
   Iman Haque \\
   Georgia Institute of Technology \\
   \texttt{ihaque3@gatech.edu} \\
}

\begin{document}

\maketitle

\begin{abstract}
  Assistive Robotics is a class of robotics concerned with aiding humans in daily care tasks that they may be inhibited from doing due to disabilities or age. While research has demonstrated that classical control methods can be used to design policies to complete these tasks, these methods can be difficult to generalize to a variety of instantiations of a task. Reinforcement learning can provide a solution to this issue, wherein robots are trained in simulation and their policies are transferred to real-world machines. In this work, we replicate a published baseline for training robots on three tasks in the Assistive Gym environment, and we explore the usage of a Recurrent Neural Network policy and Phasic Policy Gradient learning to augment the original work. Our baseline implementation meets or exceeds the baseline of the original work, however, we found that our explorations into the new methods was not as effective as we anticipated. We discuss the results of our baseline and analyze why our new methods were not as successful.
\end{abstract}

\section{Introduction}
In the United States, 27.2\% of the population lives with some form of disability, 70\% of which are considered severe \citep{b1}. These conditions vary greatly in severity and circumstance, and in many cases the patient still has a high level of independence, but may require significant assistance in conducting day-to-day tasks. Assistive Robots provide a promising solution to some of these needs.

In this project, we use the Assistive Gym framework created by \citep{b1}. The Gym includes simulation environments to train six commercially available robots to perform 6 basic assistive tasks - itch scratching, bed bathing, feeding, drinking, dressing, and arm manipulation. The original paper uses Proximal Policy Optimization (PPO) \citep{b6} to demonstrate the learning capabilities in the gym. 

We have replicated the PPO implementations of the original work for three of the tasks: feeding, itch scratching, and bed bathing. In all baseline scenarios, we achieved positive reward values, and even outperformed the original paper in one case. We then split our explorations into two orthogonal phases: replacing the MLP of the baseline with an LSTM, and replacing the PPO learning algorithm with Phasic Policy Gradient (PPG) \citep{b7}. Both of these approaches failed to improve the reward values from our baseline; the LSTM seemed unable to learn the required representation space, and the PPG implementation finished with positive but small reward values. We discuss the reasons behind the unsuitability of these approaches for the task of assistive robotics and our learnings from the same.

\section{Related work}
The Assistive Gym environment \citep{b1} provides a physics-based simulation that returns a reward according to the success of the robot in the task and the comfort preferences of the human. The simulated human can exhibit some life-like qualities that are important to studying these tasks. That is, the human can be set to be completely static, as in the case of being completely assisted by the robot, or they can be dynamic and help themselves with marginal assistance from the robot. 

The tasks which the gym simulates are based on real-world systems that have been studied in academic research. \citep{b3} provide an in-depth exploration of the feeding task, \citep{b4} conducted user-centric research by using a PR2 robot for itch scratching, and \citep{b5} developed a system to bed-bathe a patient by wiping debris off of their arms. The Assistive Gym follows the experimental setups of these papers as they have proven a level of success in achieving these tasks using real robots and patients. 

Our contributions are based on two concepts: the LSTM and Phasic Policy Gradient (PPG). The LSTM is a recurrent neural network \citep{b8} that we intend to use in place of the MLP used in the original Assistive Gym paper. PPG is a training algorithm that trains policy and value functions, benefiting from parameter sharing between the two objectives \citep{b7}.

\section{Method}

\subsection{Tasks}

We scope our work to three of the six tasks provided in the Assistive Gym: Feeding, Bed Bathing, and Itch Scratching. We paired each tasks with the robots Baxter, Sawyer, and Jaco, respectively. These robots were selected because they achieved highest reward values respectively in the original work. Images depicting examples of the training environments can be found in Figure \ref{fig:base-results}.

\subsection{Baseline}

We use the Assistive Gym as our framework of choice. At each timestep, the observations fed into policies include the 7D joint positions of the simulated robot's arm, forces applied on the robot's end effector, the effector's 3D position, 3D positions of the simulated human agent's joints and the position and orientation of the simulated human agent's head as determined by the task at hand. The policy can input actions in the environment as $\mathbb{R}^7$ changes in position of the robot's arm joints at each timestep. The framework limits forces applied on the effector to reduce the likelihood of unnatural forces being applied on the simulated human agent.

When reimplementing baselines, we use the deep reinforcement learning algorithm proximal policy optimization (PPO) to learn control policies that drive the robot's behaviour. We use the same MLP representation of two 64-dimensional hidden layers with tanh activations as the original PPO \citep{b6} and the baseline Assistive Gym \citep{b1} papers.

We created our instance of the training environment using the PPO and MLP policy implementations available from the stable-baselines3 library \citep{b12}.

Unlike \citep{b1}, where policies are trained for 10 million timesteps, we train only on 1 million timesteps. 10 timesteps make up 1 second in simulation, and each simulation rollout consists of 20 seconds. So, effectively we train our policies on 5,000 simulation rollouts for three tasks. All training is done on an Intel(R) Core(TM) i9-9900X CPU at 3.5 GHz clock speed.

Finally, we evaluate trained policies on 100 simulation rollouts and report mean rewards along with training times in the next section. We also analyze failure modes in the baseline implementation that paves way to our two contributions - using an LSTM as a memory-based deep neural network policy, and using phasic policy gradient (PPG) as our policy training algorithm.

\subsection{LSTM}

An LSTM is a type of Gated Recurrent Neural Network that uses a cell-state and hidden-state to pass "memory" information from one time step to the next \citep{b8}. This allows the modeling of sequences with long-term dependencies, that is, this model should be able to recognize if some state change early on in a sequence has an effect on a state change very late in the sequence. The LSTM is able to avoid vanishing and exploding gradients because of the presence of a forget gate along with an additive structure for cell gradients.

The intuition is that this long-term memory transduction should be able to learn the best "opening moves" in the task environment, and thereafter should maintain a context of only the actions that are useful for increasing the reward. That is, the LSTM should learn which actions move the robot toward completion of the goal the fastest. The goal of reinforcement learning is to learn a policy that yields the greatest cumulative reward, and an LSTM should be able to map early-state movements to late-state rewards.

We created an instance of the training environment from the baseline, but we replaced the 2-layer MLP with an LSTM provided by the stable-baselines library. It should be noted that this policy also uses an MLP to project the output of the LSTM to the action space.

\subsection{PPG}

Where PPG differs from PPO is that it separates policy network and value network training into two phases: the policy phase and the auxiliary phase \citep{b7}. The policy phase is essentially the same as PPO, optimizing the policy objective. The auxiliary phase optimizes an auxiliary objective, in this case the value function error. This allows the policy and value functions to share parameters and gives PPG an advantage over PPO in terms of sample efficiency and value injection into the policy. However, since more optimization processes are involved, there are more hyperparameters to deal with, and using two networks requires more memory. 

It should be noted that PPG specifies that the auxiliary objective is not limited to optimizing the value function, however it is the simplest usage of this property and is what we used. As with the previous tasks, we used an implementation of PPG available from the stable-baselines3 library and used it with the default MLP to have an orthogonal exploration.

\section{Results}

\begin{figure*}
     \centering
     \begin{subfigure}[b]{0.3\textwidth}
         \centering
         \includegraphics[width=\textwidth]{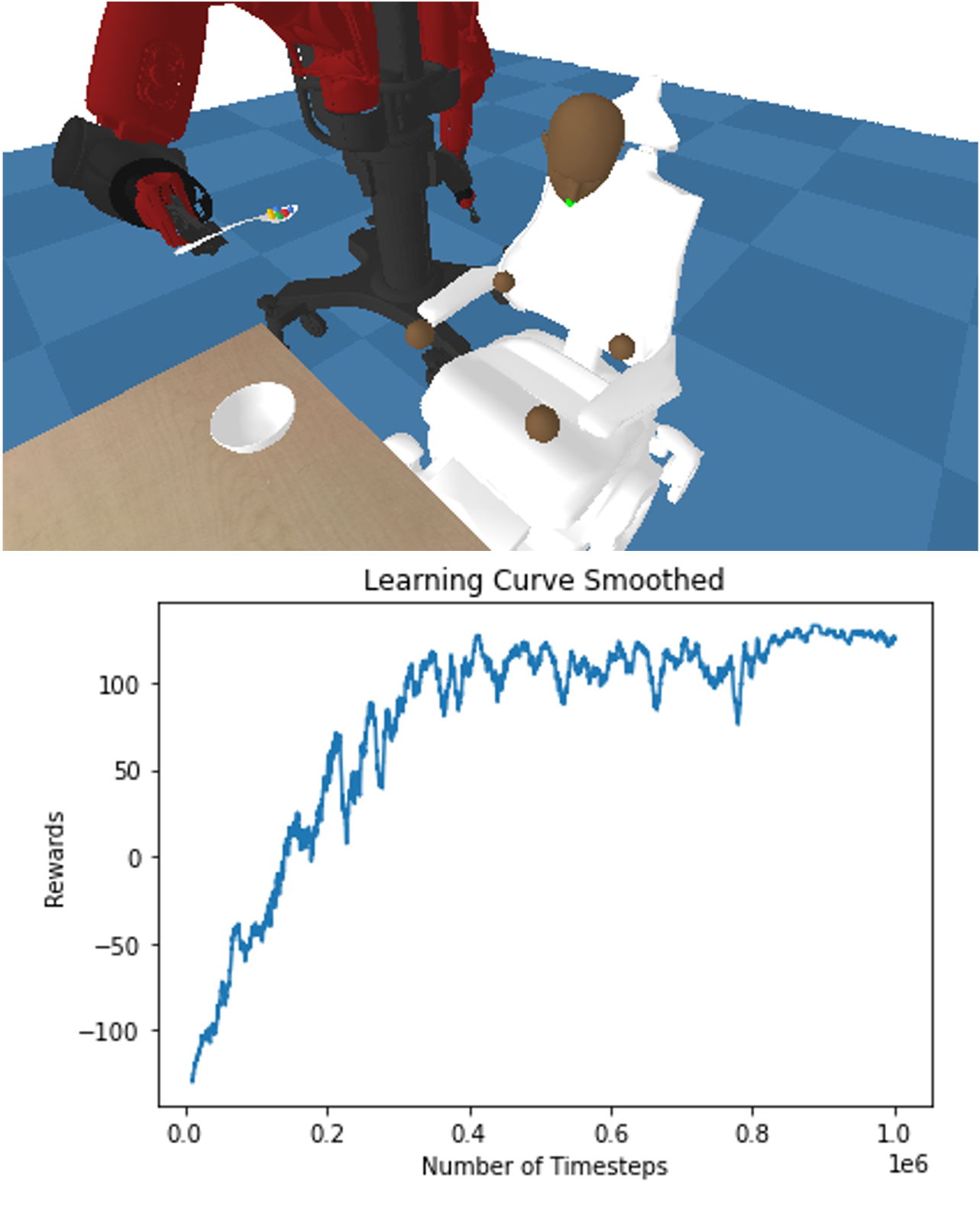}
         \caption{Feeding with Baxter}
         \label{fig:base-feeding}
     \end{subfigure}
     \hfill
     \begin{subfigure}[b]{0.3\textwidth}
         \centering
         \includegraphics[width=\textwidth]{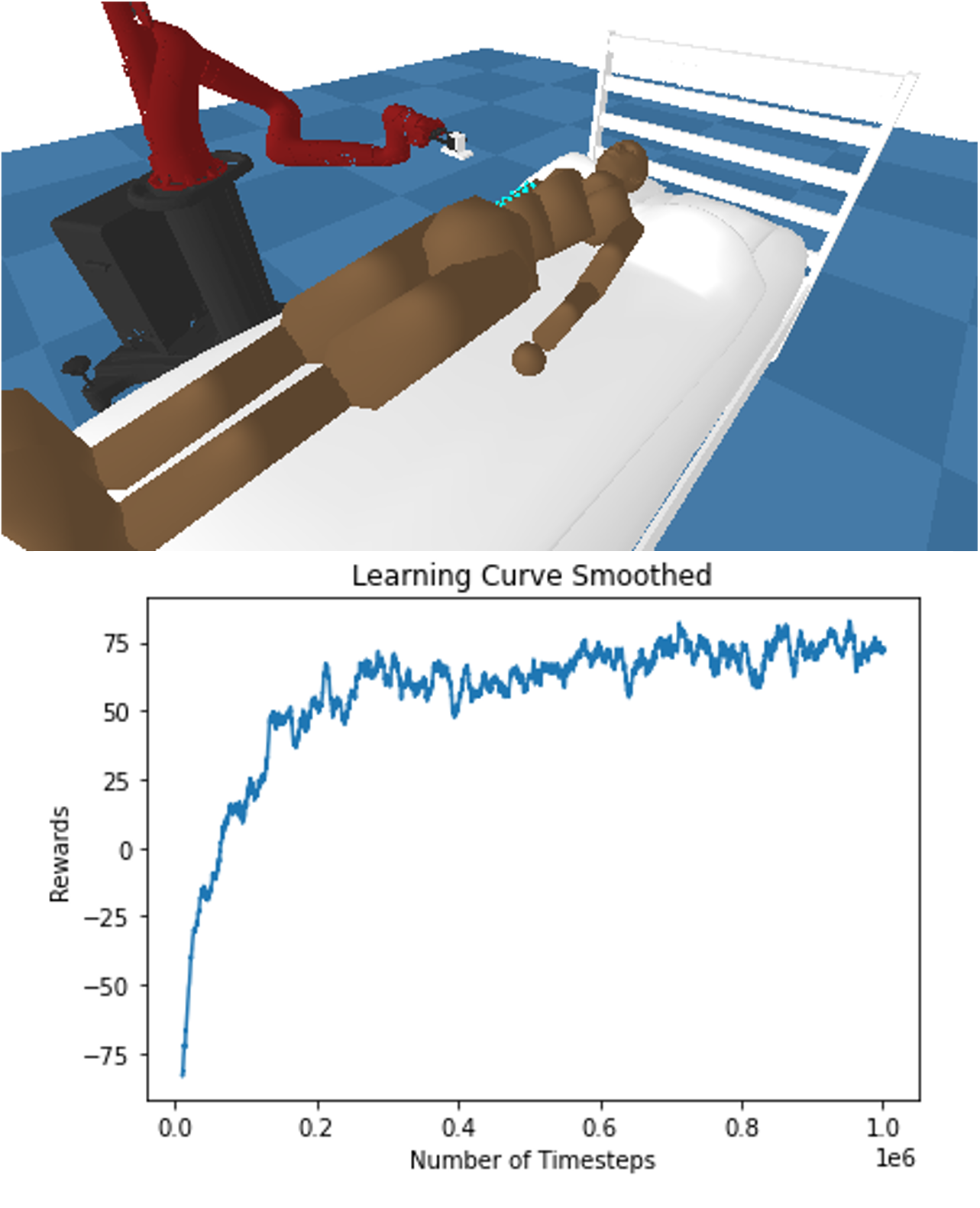}
         \caption{Bed Bathing with Sawyer}
         \label{fig:base-bathing}
     \end{subfigure}
     \hfill
     \begin{subfigure}[b]{0.3\textwidth}
         \centering
         \includegraphics[width=\textwidth]{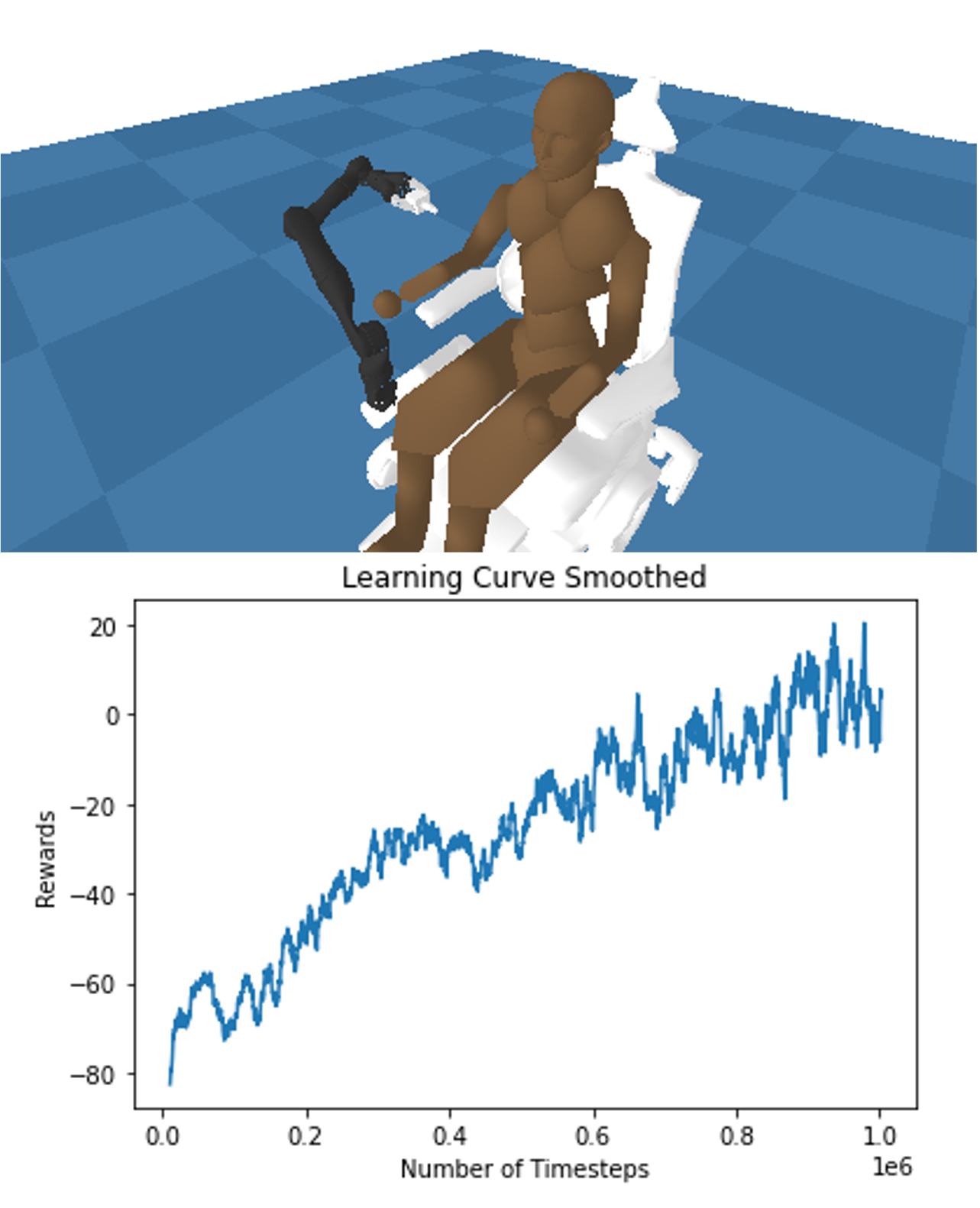}
         \caption{Itching with Jaco}
         \label{fig:base-scratch}
     \end{subfigure}
        \caption{Results of training with PPO for each of the three tasks, including a still from the simulation and the reward curve. Feeding (a) and Bed Bathing (b) exhibited saturated rewards after approximately 400k time steps. The itching task (c) showed unstable, and non-converging growth, which may indicate that the extent to which we trained was insufficient.}
        \label{fig:base-results}
\end{figure*}

Our reward values for the three tasks can be summarized in Table \ref{results_table}. The table shows a comparison between the reward results reported in the original Assistive Gym paper, and the results of our baseline reimplementation and our two contributions. Please note that we train only for 1 million timesteps as compared to \citep{b1}'s 10 million timesteps. Each of our training iterations took approximately 7-10 hours to complete. Figure \ref{fig:base-results} plots the reward curves for each of the baseline tasks and provides an image from the simulation.

The baseline reimplementation successfully produced real, positive reward values. In the case of the Feeding with Baxter task, our instance outperformed the original. While the other two tasks did not outperform the original work, we find it promising that they performed well considering the fact that they were trained on one-tenth the number of timesteps. 

\begin{figure*}
     \centering
     \begin{subfigure}[b]{0.3\textwidth}
         \centering
         \includegraphics[width=\textwidth]{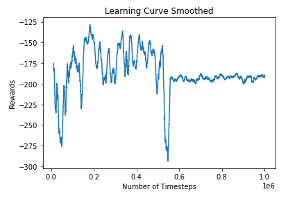}
         \caption{Feeding with Baxter}
         \label{fig:lstm-feeding}
     \end{subfigure}
     \hfill
     \begin{subfigure}[b]{0.3\textwidth}
         \centering
         \includegraphics[width=\textwidth]{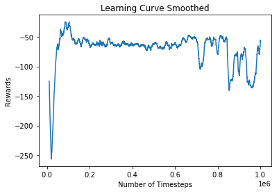}
         \caption{Bed Bathing with Sawyer}
         \label{fig:lstm-bathing}
     \end{subfigure}
     \hfill
     \begin{subfigure}[b]{0.3\textwidth}
         \centering
         \includegraphics[width=\textwidth]{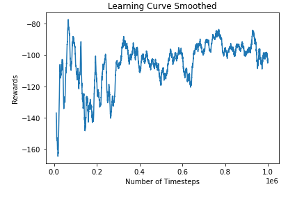}
         \caption{Itching with Jaco}
         \label{fig:lstm-scratch}
     \end{subfigure}
        \caption{Results of training with PPO using an LSTM to learn a policy for each of the three tasks. In all cases, the model was unable to learn a successful policy, indicating that the scope of the task may have been too wide for an LSTM to learn optimal reward-seeking trajectories. In the Feeding case (a) and Bed Bathing case (b), the Reward curves saturate at a negative value, after which the learning does not continue. The Itching (c) task does not saturate in this way, but it exhibits an unstable convergence at a negative reward value.}
        \label{fig:lstm-results}
\end{figure*}

The reward curves for the LSTM policies are shown in Figure \ref{fig:lstm-results}. The attempt to replace the MLP policy with an LSTM proved unsuccessful. For all tasks, the reward seemed to saturate a negative value, after which no improvement occurred. This suggests that the problem was not the time horizon - if we had trained longer it would have remained at the saturated value. 

To validate our implementation of the LSTM policy, we trained it on the 'CartPole' OpenAI gym environment with PPO. The training was successful and the result was a positive ($>200$) test time reward value. Next, we trained the policy on the 'Humanoid' environment where the task is to make a bipedal robot walk as fast as possible without falling over. The training was successful there too with a $>370$ test time reward value. Lastly, we successfully trained the policy on the 'FetchReach' environment - a closer but simpler version of our assistive robotics task, where the goal is to move a block to a predetermined position with a robotic arm.  While we made many attempts at hyperparameter tuning of the LSTM network, the policy was not able to learn the task and we got negative reward values each time. We believe the network didn't have the representation capacity to learn precise robotic arm manipulation.

\begin{figure*}
     \centering
     \begin{subfigure}[b]{0.3\textwidth}
         \centering
         \includegraphics[width=\textwidth]{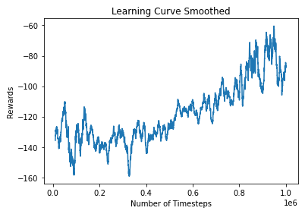}
         \caption{Feeding with Baxter}
         \label{fig:ppg-feeding}
     \end{subfigure}
     \hfill
     \begin{subfigure}[b]{0.3\textwidth}
         \centering
         \includegraphics[width=\textwidth]{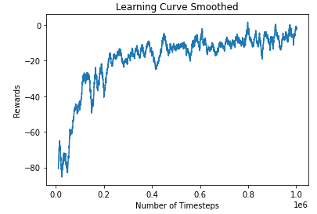}
         \caption{Bed Bathing with Sawyer}
         \label{fig:ppg-bathing}
     \end{subfigure}
     \hfill
     \begin{subfigure}[b]{0.3\textwidth}
         \centering
         \includegraphics[width=\textwidth]{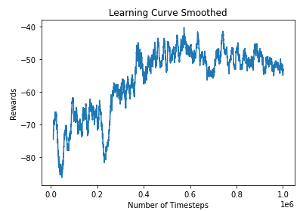}
         \caption{Itching with Jaco}
         \label{fig:ppg-scratch}
     \end{subfigure}
        \caption{Results of training with PPG using an MLP to learn a policy for each of the three tasks. PPG was able to demonstrate a more expected behavior; while the final reward values were still negative or negligibly small, they show that improvement in the rewards occurred over the 1M timesteps that the policies were trained.}
        \label{fig:ppg-results}
\end{figure*}

Finally, the reward curves for the PPG-trained policies are shown in Figure \ref{fig:ppg-results}. The usage of PPG in place of PPO returned some promising results. For all tasks, the final results are positive and the reward curves saturate. The reward value is better for one task when compared to the baseline implementation. Over the course of several training runs, we found that the training is not stable and performance may rely on the final location of the model in the reward landscape. 

\begin{table*}[t]
    \normalsize
    \caption{Comparison of reward values from the original Assistive Gym paper, our baseline implementation, and our two alphas (Baseline with LSTM, and Baseline with PPG). Our baseline implementation was able to surpass the paper's reward value on the Feeding / Baxter task. Our LSTM implementation was not able to learn a meaningful policy for any task. While PPG training did provide an acceptable training curve, }
    \label{results_table}
    \begin{center}
        \begin{tabular}{m{5em}m{4em}m{6.5em}m{4em}m{4em}}
        \hline
        \textbf{Task/ Robot}  & \textbf{Original Paper} & \textbf{Baseline \newline Implementation} & \textbf{Baseline + LSTM} & \textbf{Baseline + PPG} \\ \hline
        Feeding/ Baxter       & +108 & +114 & -110 & +130 \\ \hline
        Bed Bathing/ Sawyer   & +109 & +51  & -30  & +38  \\ \hline
        Itch Scratching/ Jaco & +281 & +46  & -89  & +11  \\ \hline
    \end{tabular}
    \end{center}
\end{table*}

\section{Discussion}

\subsection{LSTM}
In an effort to understand what went wrong with the LSTM policy, a few ideas should be considered. For one, these tasks may have been too complex to be captured by a RNN; perhaps the LSTM is not fit for a problem with these many degrees of freedom. This is validated by our 'CartPole', 'Humanoid', and 'FetchReach' tasks. On another front, does our state space lead to stale hidden states? \citep{b11} showed that "fresh" hidden states (ones calculated with current-iteration model parameters) lead to better Q-function learning than with stale hidden states (ones carried over from old model parameters). It is possible that without significant hyperparameter tuning, hidden states in the LSTM can become stale even during a single policy update. Therefore using hidden states saved in the replay buffer can become useless for predicting value functions or policy rewards. These stale hidden states may also interfere with the gradient clipping step in PPO due to their inaccuracy. They may occur because PPO does multiple gradient updates on a batch, which may affect the hidden states' distribution too greatly. As a counterexample, TRPO demonstrated that too wide a change in distribution is detrimental to policy learning.

\subsection{PPG}
Unlike PPO, where a choice has to be made between keeping policy and value networks separate or sharing a common network for both objectives, injecting learned value function features into the policy network in PPG is supposed to aid in its training while still keeping the networks separate. While, based on prior work, this is our hypothesis on why PPG should perform better than PPO, it introduces a new set of hyperparameters to optimize. We were able to improve performance for the Feeding / Baxter task, but replicating the same for other tasks seemed challenging in terms of hyperparameter tuning. Additionally, even though PPG is supposed to reduce interference between the two objectives, minimizing it requires a very specific set of hyperparameters. Our reward plots for PPG show some instability in training, which we attribute to the algorithm injecting features for two conflicting objectives in PPG. We believe more work has to be done on stabilizing PPG training, and any such further work should help this task directly.

\section{Conclusion}
To summarize, Assistive Robotics is a growing research field in which Deep Reinforcement learning can aid the improvement of patient services. We followed a published work that provided a training environment for assistive task and a performance baseline with an MLP trained using PPO. We implemented two contributions: an experiment of replacing the MLP with an LSTM, and an experiment of replacing PPO with PPG. While our baseline reimplementation was successful and outperformed the original work in one case, our LSTM experiment failed to learn a usable policy, but the PPG experiments were promising with better performance for one of the tasks. In an attempt to outperform the original work, we have demonstrated some of the inherent difficulties of training LSTM policies for arm manipulation, which should provide insights to others. As a final point, we believe that before deploying simulation-trained policies on real robots for assisting real patients, which can lead to a richer set of movements, more work needs to be done to assess the representation capacity of such policies.

{
\small
\bibliography{main}
\bibliographystyle{plainnat}
}

\end{document}